\documentclass[sigconf, nonacm]{acmart}
\usepackage{tabularx}

\usepackage{xcolor}
\usepackage{algorithm}
\usepackage{algpseudocode}
\usepackage{setspace}
\begin{document}
\title{GraphSparseNet: a Novel Method for Large Scale Traffic Flow Prediction}

%%
%% The "author" command and its associated commands are used to define the authors and their affiliations.

\author{Weiyang Kong}
\affiliation{%
  \institution{Sun Yat-Sen University}
  \city{Guangzhou}
  \state{China}
}
\email{kongwy3@mail2.sysu.edu.cn}

\author{Kaiqi Wu}

\affiliation{%
  \institution{Sun Yat-Sen University}
  \city{Guangzhou}
  \state{China}
}
\email{wukq5@mail2.sysu.edu.cn}

\author{Sen Zhang}
\affiliation{%
  \institution{Sun Yat-Sen University}
  \city{Guangzhou}
  \state{China}
}
\email{zhangs7@mail2.sysu.edu.cn}

\author{Yubao Liu}
%\orcid{0000-0001-5109-3700}
\affiliation{%
  \institution{Sun Yat-Sen University}
  \institution{Guangdong Key Laboratory of Big Data Analysis and Processing}
  \city{Guangzhou}
  \state{China}
}
\email{liuyubao@mail.sysu.edu.cn}

%%
%% The abstract is a short summary of the work to be presented in the
%% article.
\begin{abstract}

Traffic flow forecasting is a critical spatio-temporal data mining task with wide-ranging applications in intelligent route planning and dynamic traffic management. 
Recent advancements in deep learning, particularly through Graph Neural Networks (GNNs), have significantly enhanced the accuracy of these forecasts by capturing complex spatio-temporal dynamics. 
However, the scalability of GNNs remains a challenge due to their exponential growth in model complexity with increasing nodes in the graph. 
Existing methods to address this issue, including sparsification, decomposition, and kernel-based approaches, either do not fully resolve the complexity issue or risk compromising predictive accuracy. 
This paper introduces GraphSparseNet (GSNet), a novel framework designed to improve both the scalability and accuracy of GNN-based traffic forecasting models. 
GraphSparseNet is comprised of two core modules: the Feature Extractor and the Relational Compressor. 
These modules operate with linear time and space complexity, thereby reducing the overall computational complexity of the model to a linear scale.
Our extensive experiments on multiple real-world datasets demonstrate that GraphSparseNet not only significantly reduces training time by 3.51x compared to state-of-the-art linear models but also maintains high predictive performance.

\end{abstract}

\maketitle

\section{Introduction}

Traffic flow forecasting represents a quintessential spatio-temporal data mining challenge, with profound utility in a spectrum of real-world applications, including intelligent route planning, dynamic traffic management, and smart location-based services~\cite{wu2016short}.
The objective of this endeavor is to prognosticate future traffic patterns by leveraging historical traffic data, typically gleaned from the sensors embedded within transportation networks.
The advent of deep learning has revolutionized this domain, with deep learning-based techniques, particularly those grounded in graph neural networks (GNNs), emerging as a dominant force. 
GNNs excel in capturing the intricate nonlinear dynamics inherent in spatio-temporal datasets, with their efficacy underpinned by the natural alignment between traffic data structures and graph-theoretic principles \cite{Ted2022survey,jin2023survey}. 
In these models, graph nodes correspond to traffic sensors, while edges delineate the interconnections among these sensors.

Despite the predictive prowess of GNN-based methodologies, they are not without their drawbacks, most notably the exponential growth in model complexity.
The number of edges in a graph tends to increase exponentially with the number of nodes, posing a significant challenge for the scalability of GNNs to larger datasets. 
This challenge is exacerbated by the proliferation of sensors within traffic networks, driven by urban development and the pervasive integration of Internet of Things (IoT) technologies.
The vast amounts of spatio-temporal data generated by these sensors pose a challenge in applying higher-accuracy GNN methods to larger-scale datasets.

\begin{figure}
  \centering
\includegraphics[width=\linewidth]{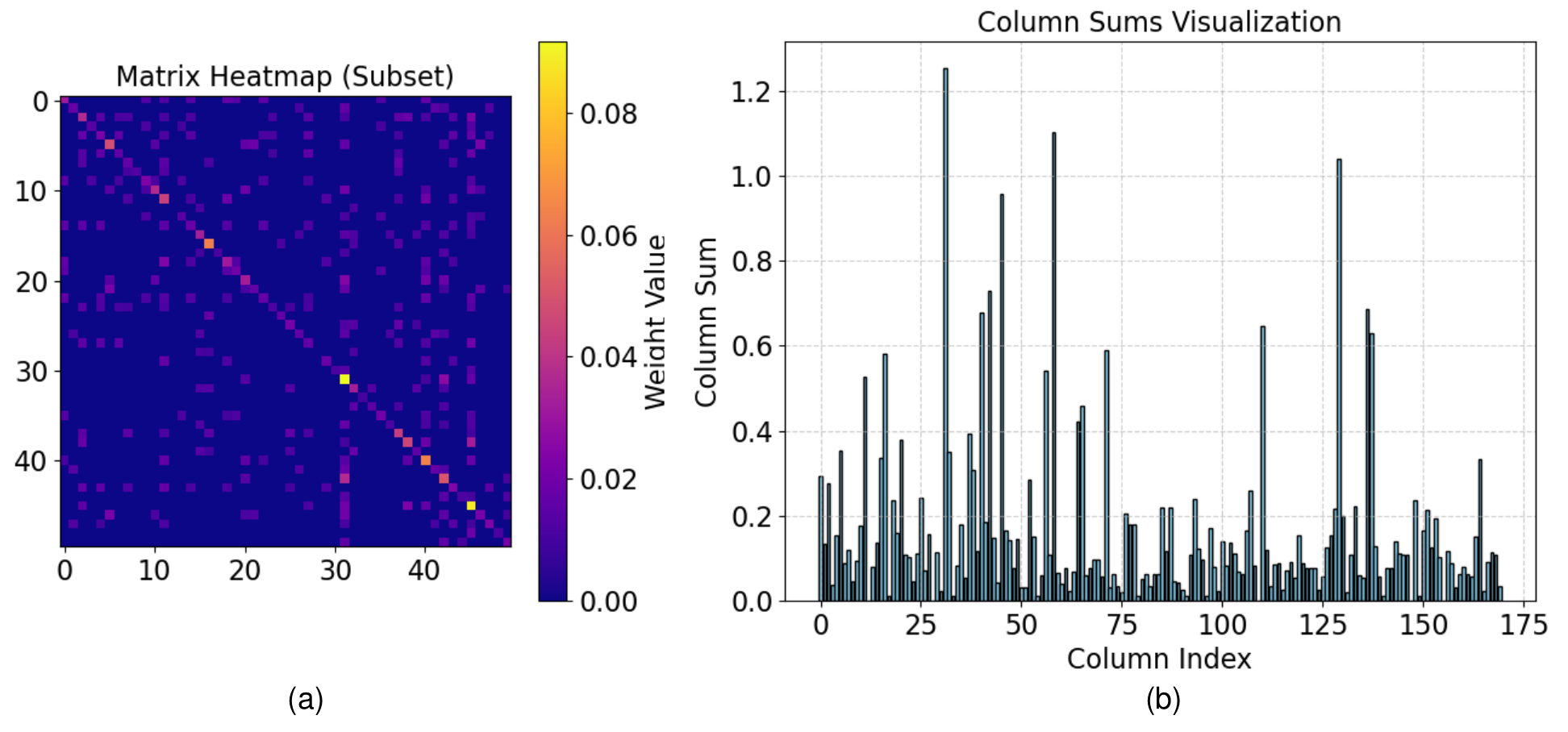}
  \caption{Visualizations correspond to the well-trained adjacency matrix obtained from existing graph neural network method. It can be observed that matrix is highly sparse. Statistical data on the weighted degree also demonstrates the sparsity of the adjacency relationships.}
  \label{fig:figure1}
\end{figure}

Several strategies have been proposed to expedite the computational efficiency of GNNs. For instance, the AGS method \cite{kdd2023ags} introduces a sparsification technique that prunes a trained model, thereby reducing its complexity during the inference phase. However, this approach still incurs substantial computational overhead during the training phase. Other methods, such as GWNet and AGCRN \cite{wu2019graph, bai2020adaptive}, employ Tucker decomposition to construct graph adjacency matrices with a reduced parameter count. Yet, these techniques merely curtail the number of trainable parameters in the adjacency matrix, without alleviating the computational burden associated with graph operations. The computational complexity of adjacency matrix multiplication, for example, remains $O(N^2)$, where $N$ denotes the number of nodes.
While some methods have successfully simplified GNNs to enhance scalability, they are not without their limitations. The BigST method \cite{bigst2024}, which employs a kernel-based approach to simplify GNNs, is susceptible to aberrant gradient values during backpropagation, potentially leading to suboptimal training outcomes and diminished predictive accuracy.

This paper aims to improve the scalability of GNNs for large-scale traffic spatio-temporal data by designing a high-precision, scalable model.
This is a non-trivial problem because simplifying GNN models faces two challenges:
First, most approaches treat accuracy and scalability in traffic prediction models as two orthogonal issues. Accurately capturing all node relationships leads to an exponential rise in computational demands. As the number of graph nodes increases, generating the adjacency matrix and performing related operations result in a sharp increase in computational complexity.
Second, existing methods for simplifying GNNs have significant limitations. Decomposition techniques fail to reduce overall model complexity, sparsification methods cannot be applied during training, and kernel-based approaches risk compromising accuracy. This highlights the need for new strategies to simplify GNNs effectively.

Upon reviewing and observing existing studies, we find that in well-trained adjacency matrices, the relationships between nodes tend to be sparse, with only a small subset of nodes being connected. 
To support this claim, we visualized a trained adaptive adjacency matrix~\cite{bai2020adaptive} and computed the weighted degree of each node, as shown in Figures 1 (a) and 1 (b). 
This trend is consistent across several models based on adaptive matrices~\cite{wu2019graph,Wu2020MTGNN,Han2021DMSTGCN}.
In Figure 1 (a), higher values in the matrix indicate stronger associations between the corresponding node pairs. 
It is evident that only a small fraction of the matrix entries have significant values, suggesting that the relationships between nodes are highly sparse.
In Figure 1 (b), we compute the weighted degree of each node, which reflects both the number and strength of connections for each node in the graph. 
As observed, with the exception of a few nodes with higher weighted degrees, most nodes maintain relatively low weighted degrees, indicating that the graph has a low connection density, with the majority of nodes having small numbers and weights of adjacent edges. 
Existing graph methods learn one entire adjacency matrix to express relationships between nodes, leading to the use of a large number of parameters and computations to learn a highly sparse matrix.
We believe that the generation of graphs and related operations can be compressed onto a much smaller space, thereby "circumventing" the computation of all node relationships in the complete graph.
We theoretically prove the rationality of this motivation, providing a theoretical derivation to show that learning one entire adjacency matrix can be replaced by learning two very small-scale matrices.

Drawing on our comprehensive analysis and key insights, we introduce a groundbreaking model, GraphSparseNet (GSNet), designed to deliver high predictive accuracy concurrently with scalability to handle large-scale datasets. 
This innovative model is comprised of two synergistic modules: the Feature Extractor, which is tasked with capturing and encoding the characteristics of graph nodes, and the Relational Compressor, which is responsible for modeling the sparse relationships between nodes.
Both modules are engineered to operate with linear time and space complexity, scaling efficiently with an increase in the number of graph nodes. This design enables the model to effectively manage and analyze large-scale data without compromising performance.
GraphSparseNet has reduced the complexity from $O(N^2)$ in most existing methods to $O(N)$. 
Even compared to the state of the art linear models, our method has improved training time by 3.51$\times$.
We evaluated the effectiveness and scalability of our proposed framework on multiple real-world datasets from different regions. 
The results of these experiments consistently demonstrate the superior performance of GraphSparseNet, achieving commendable results across all datasets evaluated.
The principal contributions of our work are encapsulated in the following points:

\begin{itemize}
\item By conducting a thorough theoretical analysis, we have identified key factors that restrict existing methods and propose innovative solutions to address these limitations.  
Our approach not only refines the predictive capabilities of GNNs but also significantly enhances their ability to scale with the increasing nodes of traffic data.  
\item We present GraphSparseNet, a novel framework specifically tailored to address the scalability challenges of large-scale traffic spatio-temporal data.  
This framework is underpinned by two core modules: the Feature Extractor and the Relational Compressor.  
The Feature Extractor capture and represent the features of each node within the graph, while the Relational Compressor innovatively models the sparse relationships between nodes. 
Both modules are designed to operate with linear time and space complexity.
\item We conducted experiments on four real-world datasets from different regions, and the results demonstrate the effectiveness and efficiency of our approach.
\end{itemize}

\section{Preliminary}

\subsection{Traffic Network}
Traffic network is represented by a undirected graph $G=(V, E)$,
where $V$ is the set of nodes (sensors), $N=\left| V \right|$ denotes the number of nodes, and $E$ is the set of edges between two nodes. 
In our problem, we assume that each node records its traffic flow data as graph signal.
A graph signal is $X^{t} \in \mathbb{R}^{N}$, where $t$ denotes the $t$-th time step.
The graph signal represents the traffic flow values at the $t$-th time step.

\subsection{Problem Definition}
Given a traffic network $G$ and its historical $S$ step graph signal matrix $X^{1:S}=(X^{1},X^{2},...,X^{S}) \in \mathbb{R}^{N \times S}$, our problem is to predict its next $T$ step graph signals, namely $X^{S+1:S+T}=(X^{S+1},X^{S+2},...,X^{S+T})$ $ \in \mathbb{R}^{N \times T}$.
We equationte the problem as finding a function $\mathcal{F}$ to forecast the next $T$ steps data based on the past $S$ steps historical data:
\begin{align}
	(X^{S+1},X^{S+2},...,X^{S+T})=\mathcal{F}((X^{1},X^{2},...,X^{S})).
\end{align}

\begin{table}[t]
	\caption{Notation}
     \renewcommand{\arraystretch}{1.1}
    \setlength{\tabcolsep}{8pt}
	\begin{tabularx}{0.5\textwidth}{p{0.08\textwidth}X}
		\toprule	
		\textbf{Notation} & \textbf{Meaning}\\
  \hline
        $G$ & The undirected graph.\\
        $V$ & The set of nodes (sensors).\\
        $E$ & The set of edges between nods.\\
        $N$ & The number of nodes.\\
        $A$ & The adjacency matrix of graph $G$.\\
        $X$ & The graph signal.\\
        $S$ & The number of historical time step of $X$.\\
        $T$ & The number of predict time step of output.\\
        $H$ & The hidden states in model.\\
        $C$ & The dimension of the compressed space.\\
        $K$ & The adjacency matrix in low-dimensional space.\\
        $U$ & The coefficient matrix in low-dimensional space.\\
        $W,B$ & The trainable parameter of the pivotal graph convolution.\\
        $V$ & The transformation matrix.\\
		\bottomrule
	\end{tabularx}
\end{table}

\section{Analysis}
In this section, we discuss the constraints that hinder the scalability of graph neural network methods.
We also explain our motivations for simplifying models, supported by theoretical evidence and empirical findings.

Graph neural network methodologies are highly effective in traffic prediction, as they adeptly capture node features while facilitating the integration of features across different nodes in accordance with the graph's topology. 
The most prevalent graph neural networks can be represented in the following form:
\begin{equation}
  H = Agg(A,X,\Theta)
\end{equation}
Here, $H$ represents the hidden state of the model, $A$ denotes the adjacency matrix, $X$ represents the input data, and $\Theta$ refers to the trainable parameters. 
The function $Agg(\cdot)$ is a specific method for aggregating elements in $X$ based on the relationships defined in $A$. 
A commonly employed approach for this aggregation function is the use of spatial domain graph convolution operations in the following form:
\begin{equation}
  H = \sigma(AXW)
\end{equation}
Here, $\sigma(\cdot)$ represents arbitrary activation function, and $W$ denotes the trainable parameters.
It leverage matrix multiplication to integrate input data $X$ in accordance with the relationships specified by the adjacency matrix $A$.
In the majority of current research\cite{wu2019graph, bai2020adaptive}, to achieve superior predictive performance, matrix 
$A$ is often composed of trainable parameters, a practice referred to as the adaptive adjacency matrix. 
This approach allows the model to reduce its reliance on prior knowledge while enhancing its representational capacity for potential node associations. 
To prevent over-fitting due to the extensive use of parameters, this adaptive matrix is typically factorized into the product of two smaller matrices:
\begin{equation}
  A = SoftMax(E_1E_2)
\end{equation}
Here, $E_1 \in  \mathbb{R}^{N \times C}$ and $E_2 \in  \mathbb{R}^{C \times N}$ are both trainable parameters, $C$ is a hyperparameter.
This technique reduces the number of parameters required to construct the adjacency matrix from $N^2$ to $2CN$
Employing this technique, a series of methods have empirically demonstrated that the effectiveness of the adaptive adjacency matrix can be maintained even when the parameter $C$ is set to a value significantly smaller than $N$.
When $N$ is sufficiently large, employing a decomposition approach can significantly reduce the number of training parameters needed to construct the adjacency matrix $A$.
However, the complexity of the matrix multiplication for $AX$ in graph convolution is still $O(N^2)$, which greatly limits the scalability of this operation on large-scale data.

AGS~\cite{kdd2023ags} proposed the use of a mask matrix to simplify the adjacency matrix $A$. 
This approach introduces a Boolean mask matrix $\beta$, which has the same dimensions as the adjacency matrix $A$, after $A$ has been trained.
By performing the Hadamard product of $A$ and $\beta$, the final adjacency matrix is determined based on the binary gating within $\beta$, deciding whether to retain the connections. 
However, this method can only theoretically reduce the operational speed during the model's inference phase, as it requires the retraining of $\beta$ to achieve a sparse adjacency matrix after $A$ has been trained. 
Simultaneously training $A$ and $\beta$ can lead to difficulties in model convergence, thereby diminishing the model's predictive accuracy.

BigST~\cite{bigst2024} designed a graph convolutional model with linear complexity.
However, BigST can lead to the generation of anomalous gradient values during the model's training process, which affects the training effectiveness and the ultimate predictive accuracy.
For graph convolution operations under normal circumstances, The Equation 4 can be rewritten as
\begin{equation}
  A_{ij} = \frac{f(s_{ij})}{\sum_{k=1}^{N} f(s_{ik})}
\end{equation}
\begin{equation}
  s_{ij} = E_{1i}E_{2j},~f(x) = exp(x)
\end{equation}
The gradients of the matrix $A$ is derived as~\cite{kerneldevil}:
\begin{equation}
  \frac{\partial A_{ij}}{\partial s_{ik}} = \frac{f'(s_{ik})}{f(s_{ik})} (1_{j=k}A_{ij}-A_{ij}A_{ik})
\end{equation}
If Equation (6) is substituted into Equation (7), we can deduce that:
\begin{equation}
 \lvert  \frac{\partial A_{ij}}{\partial s_{ik}} \rvert \leq \frac{1}{4}
\end{equation}
It is evident that the gradients of the parameters do not exhibit any significant anomalies.
BigST employs a kernel function mapping approach to simplify the graph convolution operation into a linear product of multiple matrices, as follows:
\begin{equation}
  s_{ij} = \phi(E_{1i})\phi(E_{2j}),~f(x) = x
\end{equation}
Here $\phi(\cdot)$ is the kernel function.
If Equation 9 is substituted into Equation 7, we can deduce that:
\begin{equation}
 \lvert  \frac{\partial A_{ij}}{\partial s_{ik}} \rvert \leq \frac{1}{4 \lvert s_{ik}\rvert}
\end{equation}
There is always a probability that the gradients during back propagation will produce anomalous values since $s_{ij}$ is set from 0 to positive infinity in the BigST.

Our goal is to approximate the graph convolution operation in Equation 3 with a computational complexity of $O(N)$, while maintaining the model's accuracy. 
This is feasible because, as shown in Figure~\ref{fig:figure1}, well-trained adaptive matrices tend to be highly sparse.
For the factorized adjacency matrix $A = SoftMax(E_1 E_2)$, we consider the properties of matrices $E_1$ and $E_2$ that: 
\begin{equation}
  Rank(E_1) \leq C
\end{equation}

\begin{equation}
  Rank(E_2) \leq C
\end{equation}

\begin{equation}
  Rank(E_1E_2) \leq Min(Rank(E_1), Rank(E_2)) \leq C
\end{equation}
This implies that the column (and row) space of the matrix $E_1E_2$ can both be spanned by a set of $C$ linearly independent vectors. 
In other words, we have the following theorem:

\begin{theorem}
Let $M \in \mathbb{R}^{N \times N}$ be a matrix with rank $C$. 
There always exists a non-unique matrix $K \in \mathbb{R}^{C \times C}$ such that matrix $M$ can be constructed via some matrix multiplication transformations involving $K$.
\end{theorem}

\textbf{Proof:} 

For a matrix $M$ of rank $C$, it is known from matrix theory that we can factorize $A$ as a product of three matrices:
\begin{equation}
M = \lambda \times K \times \mu
\end{equation}
Here, the matrix $\lambda \in \mathbb{R}^{N \times C}$ consists of a set of linearly independent column vectors from the column space of $M$. 
The matrix $\mu \in \mathbb{R}^{C \times N}$ consists of a set of linearly independent row vectors from the row space of $M$. 
Matrix $K$ captures the transformation between these basis vectors.
In this factorization, $\lambda$ and $\mu$ are matrices formed from linearly independent column and row vectors (basis vectors) from the column space and row space of $M$, respectively. The matrix $K$ represents the transformation between these basis vectors.

To find the matrices $\lambda$ , $\mu$ and $K$, the following steps are taken:
Perform column operations on matrix $M$ (e.g., using Gaussian elimination) to identify a set of linearly independent column vectors that span the column space of $M$.
These basis vectors form the columns of matrix $\lambda$. 
Similarly, perform row operations on matrix $M$ to form the rows of matrix $\mu$. 

Matrix $K$ is determined by the linear combination relationships between the basis vectors in $\lambda$ and $\mu$. 
According to the factorization, matrix $M$ can be written as $M = \lambda \times K \times \mu$. 
In other words, every element of matrix $M$ is obtained by a linear combination of the column basis vectors in $\lambda$ and the row basis vectors in $\mu$, with the coefficients coming from matrix $K$.
Write each column of matrix $M$ as a linear combination of the column basis vectors in $\lambda$, with the coefficients being the elements of matrix $K$. 
Similarly, write each row of matrix $M$ as a linear combination of the row basis vectors in $\mu$, with the coefficients being the elements of matrix $K$.
By comparing each element of matrix $M$ with the corresponding linear combinations of basis vectors, a system of linear equations can be formed. 
Solving this system gives the elements of matrix $K$. 
Specifically, for each element $M_{ij}$ of matrix $M$, we can write the following equation:
\begin{equation}
M_{ij} = \sum_{k=1}^{C} \lambda_{ik} \cdot K_{kl} \cdot \mu_{lj}
\end{equation}
Solving these equations yields the elements of matrix $K$.

Next, we will prove that K is not unique.
The matrices $\lambda$ and $\mu$ are not uniquely determined because there are multiple ways to choose the basis vectors for the column space and row space of $M$. 
Consequently, there may be different choices of $K$.
The matrix $\lambda$ is formed by selecting a basis for the column space of $M$, but there can be multiple different sets of linearly independent column vectors that span the same column space. Thus, $\lambda$ is not unique.
Similarly, $\mu$ is not unique.
Since both $\lambda$ and $\mu$ can vary, matrix $K$ is not uniquely determined.
We shall further elucidate the aforementioned properties through the construction of an illustrative example.
Suppose we have a factorization $M = \lambda \times K \times \mu$. 
Now, consider applying an invertible linear transformation to the basis vectors. 
Specifically, let $D$ be an arbitrary invertible $C \times C$ matrix. 
Define new matrices $\tilde{\lambda}$ and $\tilde{\mu}$ as:
\begin{equation}
\tilde{\lambda} = \lambda \times D
\quad \text{and} \quad
\tilde{\mu} = D^{-1} \times \mu
\end{equation}
Then, we can construct a new factorization:
\begin{equation}
M = \tilde{\lambda} \times (D^{-1} \times K \times D) \times \tilde{\mu}
\end{equation}
In this new factorization, the matrix $\tilde{K} = D^{-1} \times K \times D$ differs from the original matrix $K$. 
Hence, by applying different invertible transformations, we can generate different matrices $K$ that still satisfy the equation $M = \tilde{\lambda} \times \tilde{K} \times \tilde{\mu}$.
This demonstrates that matrix $K$ is not unique, as different transformations of the basis vectors lead to different but valid choices for $K$.
\hfill \(\square\)

For the sake of convenience in subsequent descriptions, we assume, based on the aforementioned theorem, that:
\begin{equation}
  E_{1}E_{2}^{ij} = \sum_{m=1}^{C} \sum_{n=1}^{C} U_{mn}^{ij}K_{mn} ,
\end{equation}
Here $U$ represents the coefficient of the linear combination, and $K_{mn}$ denotes the elements of matrix $K$ with rank $C$.
Thus, we can consider that matrix $E_{1}E_{2}$ can be derived from matrix $K$ and coefficient matrix $U$ through a series of linear transformations. 
In other words, the process of the model learning matrix $E_{1}E_{2}$ is equivalent to learning matrix $K$ and the coefficient matrix $U$.

Since the well-trained adjacency matrix $A$ is sparse, we propose that a significant number of values in $U$ can be replaced with a fixed number, meaning that the number of trainable parameters used to express the combination coefficients in matrix $U$ can be much smaller than its shape.
The adjacency matrix $A$ used in existing model is obtained by applying the activation function to $E_{1}E_{2}$. 
Taking the most commonly used form, $SoftMax()$, as an example, we have:
\begin{equation}
  SoftMax(E_{1}E_{2}^{ij}) = \prod_{i=1}^{C} \prod_{j=1}^{C} SoftMax(K_{ij})^{u_{ij}}
\end{equation}
If $A_{ij}$ in the adjacency matrix is zero, we only need to set the corresponding coefficient in $U$ to a fixed negative value (for example, negative 3) to ensure that the calculated $A_{ij}$ is close to zero. 
For a sparse adjacency matrix $A$, most elements in $U$ would be set to negative 3, with only a small portion of elements needing to be determined through training. 
The proportion of trainable parameters in matrix $U$ is positively correlated with the sparsity of matrix $A$. 
Similarly, if other activation functions are chosen, we simply need to set the corresponding elements in $U$ to other fixed values since the computation of the activation function is fixed.

Based on the aforementioned analysis, we can consider that in the adaptive matrix based methods, the technique of setting $A$ as a trainable parameter can be replaced by allowing the model to learn matrix $K$ and the linear combination coefficient matrix $U$. 
Intuitively, the information contained in matrix $A$ is equivalent to the information contained in $K$ and $U$.

Furthermore, in the graph convolution operation of equation 3, the intention of the matrix multiplication for $AX$ is to perform feature fusion of the input data $X$ according to the adjacency relationship $A$ through linear transformation. 
The complexity of this operation is $O(N^2)$, which limits the scalability of the graph convolution operation.
Since we can learn matrix $K$ and the combination coefficients $U$ separately, in this paper, we propose a new framework to replace the graph convolution operation in equation 3. 
The following sections will introduce the specific details of our newly proposed method.

\section{Proposed Model}
we commence by elucidating the overarching framework of the model. Subsequently, we introduce the specific module designs within the model.

\begin{figure*}
  \centering
  \includegraphics[width=\linewidth]{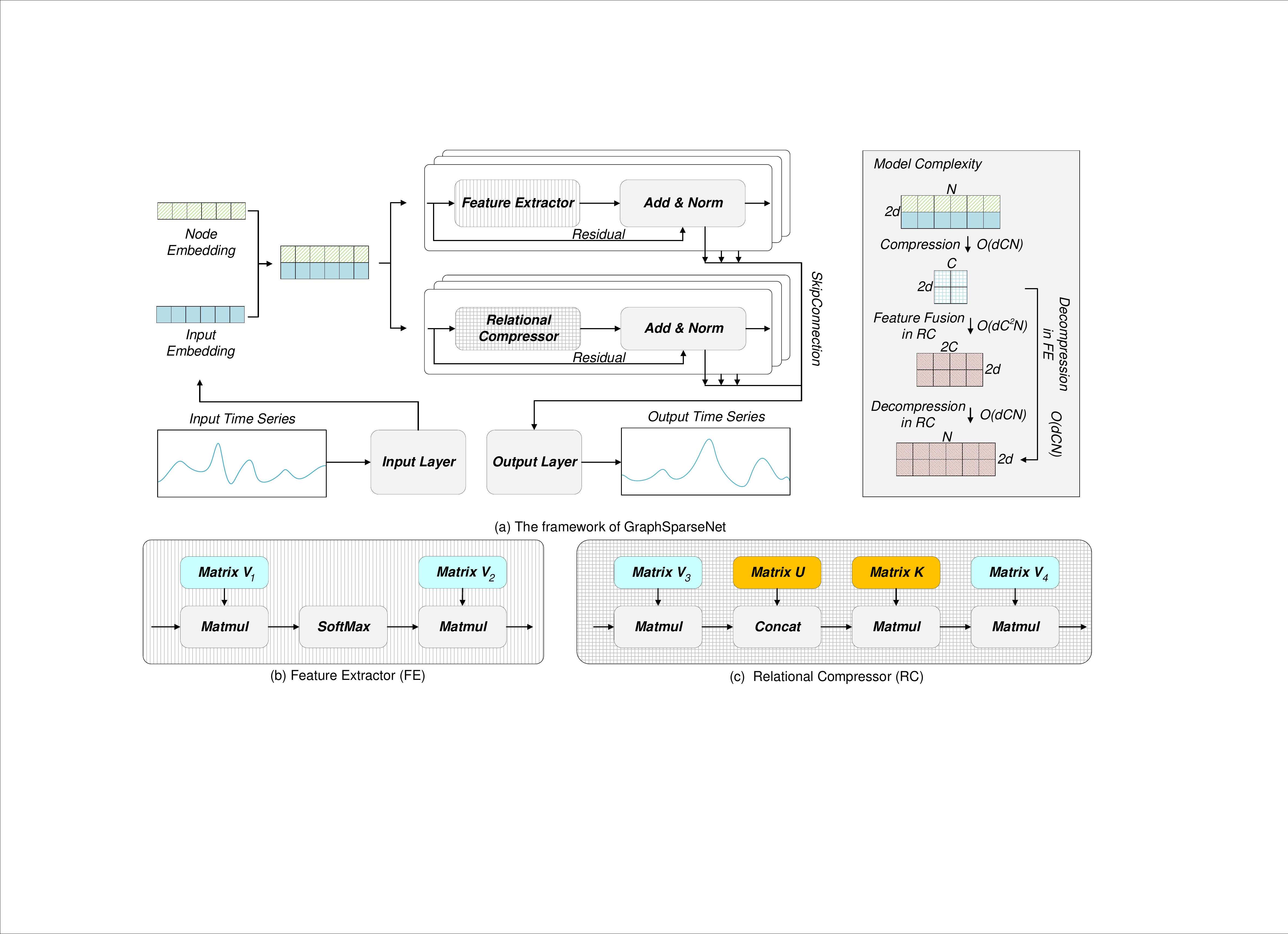}
  \caption{An illustration of the framework.}
  \label{fig:framework}
\end{figure*}

\subsection{Framework of the model}

As depicted in Figure~\ref{fig:framework}, our model comprises two distinct modules: Feature Extractor and Relational Compressor. 
The Feature Extractor compresses the input data into an low-dimensional space and subsequently reconstructs it, while simultaneously updating an embedding that represents the node features. 
The Relational Compressor transforms the input data into an low-dimensional space, models the implicit adjacency relationships between different nodes in the low-dimensional space, and ultimately reconstructs the data back into its original space. 
The low-dimensional spaces involved in the two modules are mutually aligned.
Both the Feature Extractor and Relational Compressor are stacked in a serial manner, and the outputs from each layer are concatenated with skip connections, which are then transformed into prediction through an output layer.

In our model, there are two types of embedding matrices.
One is the input embedding $P \in \mathbb{R}^{N \times d}$, and the other is the node embedding $Q \in \mathbb{R}^{N \times d}$.
$d$ denotes the number of channel in the embedding.
The node embedding is composed of trainable parameters, the elements of which are updated exclusively during the training process of the model and do not vary with the input data. In contrast, the input embedding is generated from the model's input data and automatically adjusts according to different inputs.
The purpose of the node embedding is to model the characteristics of each node in the transportation network during the training process of the model.
The input embedding encompasses not only the node feature information but also the spatial information (such as the relationship between nodes) and temporal information present in the input data.
The objectives of transforming the input data into input embeddings are twofold: first, to obtain a dense representation for spatio-temporal features, and second, to enable the model to automatically learn the most salient features of the data.

\subsection{Feature Extractor}
%Feature Extractor and Relational Compressor.

The advantages of Graph Neural Networks stem from two aspects: one is their powerful feature learning capability, which can accurately capture the features of each node, and the other is their ability to handle graph-structured data, allowing node features to be integrated based on adjacency relationships. 
By aggregating information from neighboring nodes to update the representation of a node, it can capture complex relationships between nodes, thereby better understanding graph-structured data.
In the Feature Extractor module, our primary goal is to enable the model to learn the features of different nodes.

In the design of the Feature Extractor, we have two objectives: one is to update the node embeddings through this module to model the characteristics of each node in the traffic data, and the other is to attempt to compress the input embeddings, which contain spatio-temporal information, to a low-dimensional space, thereby preparing for subsequent feature fusion in low-dimensional space within the model.

To achieve both of these objectives, we have designed a module that compresses and then decompresses the input information based on node features.
The input embeddings encompass only the local spatio-temporal characteristics of the input time series, as the data presented to the model at each instance constitutes a small slice of the entire dataset. Concurrently, the node embeddings remain invariant over time, thereby capturing and representing global information.
Therefore, we concatenate the two types of embeddings as the input to the module. 
\begin{equation}
  X_{FE} = P||Q
\end{equation}
Here, $X_{FE}$ represents the input of Feature Extractor.
The concatenated input $X_{FE}$ contains both the local spatio-temporal information of the input data and the global features of each node.
A reasonable approach to compressing data from an $N$-dimensional space to a low-dimensional space is to cluster based on node features. 
To this end, we use a matrix generated from the node features to compress the module input, as follows:
\begin{equation}
  H_{FE} = V_1X_{FE}
\end{equation}
\begin{equation}
  V_1 = W_1Q + B_1
\end{equation}
Here, $H_{FE}$ represents the hidden state, $V_1$ is the compressing matrix.
$W_1$ and $B_1$ are trainable parameter matrices.
Subsequently, we designed the decompression process, which is similar to the compression.
To introduce non-linear characteristics and control in the Feature Extractor, we added an activation function between the compression and decompression processes. 
We utilize the SoftMax function to present the hidden layer in the form of a probability distribution, thereby enhancing the training effectiveness of this module during the decompression process.
\begin{equation}
  O_{FE} = V_2SoftMax(H_{FE})
\end{equation}
\begin{equation}
  V_2 = W_2Q + B_2
\end{equation}

%----

On one hand, the process of compression and decompression can update the a node embedding $Q$ to learn the features of different nodes. 
The information contained within this embedding will be shared with Relational Compressor.
On the other hand, the model is designed to learn a method of compressing the input to a size of $C$ dimensions, which serves as an auxiliary function for Relational Compressor to map the input into a low-dimensional space. 
This alignment of the compression ratio in this module with that in Relational Compressor is crucial for the seamless integration and effective operation of the overall model.

\begin{figure}
  \centering
  \includegraphics[width=0.8\linewidth]{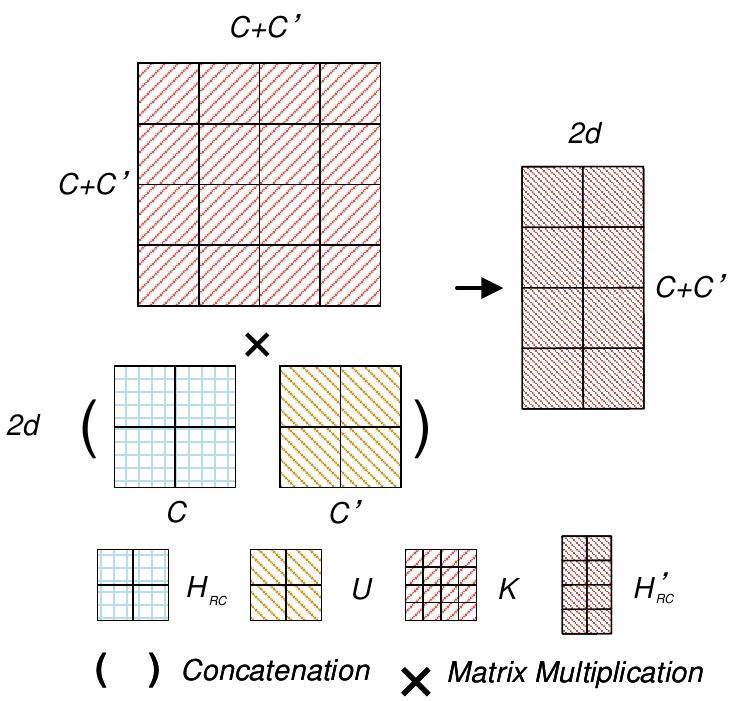}
  \caption{An illustration in Relational Compressor.}
  \label{fig:rc}
\end{figure}

\subsection{Relational Compressor}

%----

As we discussed in the previous analysis, learning the adjacency matrix $A$ can be equivalent to learning matrices $K$ and $U$. 
In the design of the Relational Compressor, our objective is to learn the adjacency matrix $K$ and the coefficient matrix $U$ of the traffic graph in the low-dimensional space, and to achieve the fusion of node features in the low-dimensional space to avoid the $O(N^2)$ complexity in spatial graph convolutions.
To this end, we also design the model with a compression and decompression manner, which is similar to Feature Extractor.
Consistent with the concept in the Feature Extractor, this module also takes the concatenation of node embeddings and input embeddings as input:
\begin{equation}
H_{RC} = V_3(P || Q)
\end{equation}
Here $H_{RC}$ is the hidden state in Relational Compressor.
Unlike the Feature Extractor, in the compression process, to enable the compression to consider the local spatio-temporal features in the input embedding $P$, rather than compressing solely based on global node features, we consider using two embeddings to generate the compression matrix:
\begin{equation}
V_3 = W_3(P || Q) + B_3
\end{equation}
After compressing the input into the low-dimensional space, we let the model learn the coefficient matrix $U$ in the hidden state through concatenation.
\begin{equation}
H_{RC}' = H_{RC}||U
\end{equation}
As shown in Figure~\ref{fig:rc}, the shape of the coefficient matrix $U$ can be controlled by $C'$.
Subsequently, we fuse features according to the low-dimensional adaptive adjacency matrix $K$ through matrix multiplication in the hidden state, as follows:
\begin{equation}
H_{RC}'' = KH_{RC}'
\end{equation}
$K$ is the adaptive adjacency matrix in the $C+C'$-dimensional space.
Both $K$ and $U$ are composed of trainable parameters, and in Section 2, we discussed the feasibility of using $K$ and $U$ to replace the adaptive matrix $A$. 
Through concatenation and matrix multiplication, Relational Compressor can achieve the purpose of learning adjacency relationships and feature fusion in the low-dimensional space.
The number of trainable parameters contained in matrices $K$ and $U$ is far less than that in $A$.
Moreover, the computational complexity of the module is also far less than that of spatial GCN and its existing improved schemes, which we will analyze in detail later.
Subsequently, we architected the decompression process to be analogous to the compression process, ensuring a symmetrical approach to feature representation.
\begin{equation}
O_{RC} = V_4SoftMax(H_{RC}'')
\end{equation}
\begin{equation}
V_4 = W_4(P || Q) + B_4
\end{equation}

The outputs of the two modules are summarized through a skip-connection approach and the final prediction result is obtained through the output layer.
The final output is:
\begin{equation}
O = O_{FE} \parallel O_{RC}
\end{equation}
The output layer consists of an activation function and a fully connected layer.

The overall training process of our model is outlined in Algorithm~\ref{alg1}.

\begin{algorithm}
\caption{Training Schema of GraphSparseNet} 
\label{alg1} 
\setstretch{1.15}
\begin{algorithmic}[1]
\State \textbf{Input:} Graph $G$, Graph signal $X$
\State \textbf{Output:} Model prediction $\hat{X}$
\State
\State $P \gets$ InputLayer($X$) \Comment{Transform $X$ to input embedding $P$}
\State $Q \gets$ NodeEmbedding($G$) \Comment{Obtain node embeddings $Q$}
\State
\State $V_1, V_2 \gets$ GenerateTransformationMatrices($Q$) \Comment{Generate transformation matrices}
\State $V_3, V_4 \gets$ GenerateTransformationMatrices($P || Q$) \Comment{Generate transformation matrices from concatenation}
\State
\State \textbf{Modules:} FE (Feature Extractor), RC (Relational Compressor)
\For{$i \gets 1$ to L} \Comment{Loop for L layers}
    \State \textbf{FE Module:}
    \State $H_{FE} \gets V_1 \cdot (P || Q)$ \Comment{Compress the concatenation with $V_1$}
    \State $H_{FE} \gets$ ActivationFunction($H_{FE}$) \Comment{Apply activation function}
    \State $H_{FE} \gets V_2 \cdot H_{FE}$ \Comment{Decompress with $V_2$}
    \State $O_{FE} \gets$ ResidualConnection($P || Q, H_{FE}$) \Comment{Residual connection and normalization}
    \State \textbf{RC Module:}
    \State $H_{RC} \gets V_3 \cdot (P || Q)$ \Comment{Compress the concatenation with $V_3$}
    \State $H_{RC}' \gets H_{RC} || U$ \Comment{Concatenate with matrix $U$}
    \State $H_{RC}'' \gets K \cdot H_{RC}'$ \Comment{Transform with matrix $K$}
    \State $H_{RC}'' \gets$ ActivationFunction($H_{RC}''$) \Comment{Apply activation function}
    \State $H_{RC}'' \gets V_4 \cdot H_{RC}''$ \Comment{Decompress with $V_4$}
    \State $O_{RC} \gets$ ResidualConnection($P || Q, H_{RC}''$) \Comment{Residual connection and normalization}
    \State $O \gets$ UpdateInput($O_{FE}, O_{RC}$) \Comment{Update input for next layer}
\EndFor
\State $Output \gets$ FullyConnectedLayer($O$) \Comment{Final fully connected layer}
\State $\hat{X} \gets$ ActivationFunction($Output$) \Comment{Apply activation function for prediction}
\State \textbf{Backpropagation:} Compare($\hat{X}, X_{\text{true}}$) \Comment{Compare prediction with true value}
\end{algorithmic}
\end{algorithm}

\subsection{Complexity Analysis}
In this section, we analyze the complexity of various modules within the proposed model.
For the sake of simplicity, we assume that 
$C = C' = d$ in the subsequent discussions.

Regarding time complexity:
The complexity of converting the input graph signal to input embeddings in the input layer is $O(NSC)$, where $N$ is the number of nodes, $S$ is the number of historical time steps, and $C$ is the dimensionality of the low-dimensional space.
In the Feature Extractor, the complexity of compression and decompression is $O(C^2N + C^2N)$.
In the Relational Compressor, the complexity of compression and decompression is $O(C^2N + C^2N)$, and the complexity of feature fusion in the low-dimensional space is $O(C^3)$.
The complexity of the output layer is $O(NTC)$, where $T$ is the time step of the output graph signal.
Since $C$, $S$, and $T$ are constants, the overall time complexity of the model is $O(N)$.

Regarding space complexity:
The space complexity for input embeddings and node embeddings is $O(SN + CN)$.
The space complexity for generating the compression and decompression transformation matrices is $O(CN)$.
The space complexity for generating $K$ and $U$ is $O(CN + C^2)$.
The overall space complexity of the model is $O(N)$.

It can be observed that our approach exhibits a space complexity and time complexity of $O(N)$, which endows our method with superior scalability.

\section{Experments}

In this section, we evaluated our proposed model by empirically examining on four real-world datasets with the state-of-the-art models for traffic forecasting. 
To support the reproducibility of the results in this paper, we have released our code on website. \footnote{https://github.com/PolynomeK/GSNet}

\begin{table}[t]

  \centering
    \caption{Datasets description.}
    \renewcommand{\arraystretch}{1.1}
    \setlength{\tabcolsep}{6pt}
    %\resizebox{\linewidth}{!}{
    \begin{tabular}{ccccccccc}
    \hline
    Dataset & Node  & Samples & Sample Rate & Traffic Record  \\
    \hline
    PEMS08 & 170   & 17,856 & 5min  &  3,035,520 \\
    England & 314   & 17,353 & 15min  &  5,448,842 \\
    PEMS07 & 883   & 28,224 & 5min & 24,921,792   \\
    CA & 8,600  & 17,280  & 5min   & 148,608,000 \\
    \hline
    \end{tabular}%
    %}

  \label{tab:properties}
\end{table}

\subsection{Datasets and Pre-processing}

We conduct experiments on four widely used real-world public traffic datasets from: (1)PEMS\footnote{http://pems.dot.ca.gov/}(PEMS07 and PEMS08), (2)England\footnote{http://tris.highwaysengland.co.uk/detail/trafficflowdata}, and (3)CA\cite{liu2023largest}.
The biggest dataset is California (CA), including a total number of 8,600 sensors. 
To the best of our knowledge, CA is currently the publicly available dataset with the highest number of nodes recorded by loop detectors.
It contain three representative areas: Greater Los Angeles, Greater Bay Area, and San Diego.
A brief description are given in Table~\ref{tab:properties}.

Z-score normalization is applied to inputs as follows.
\begin{equation}
	X_{input} = \frac{X - mean(X)}{std(X)}
\end{equation}
Here $mean(X)$ and $std(X)$ denote the mean and the standard deviation of the historical time series, respectively.

% Table generated by Excel2LaTeX from sheet 'Sheet1'
\begin{table*}[htbp]
  \centering
\caption{  Performance comparison of different approaches on all the datasets.}
  \renewcommand{\arraystretch}{1.1}
  \setlength{\tabcolsep}{6.5pt}
    \begin{tabular}{l|ccc|ccc|ccc|ccc}
    \hline
    Dataset & \multicolumn{3}{c|}{PEMS08}\ & \multicolumn{3}{c|}{England}    & \multicolumn{3}{c|}{PEMS07}  & \multicolumn{3}{c}{CA}  \\
    \hline
    Metric & \multicolumn{1}{c}{MAE} & \multicolumn{1}{c}{MAPE} & \multicolumn{1}{c|}{RMSE} & \multicolumn{1}{c}{MAE} & \multicolumn{1}{c}{MAPE} & \multicolumn{1}{c|}{RMSE} & \multicolumn{1}{c}{MAE} & \multicolumn{1}{c}{MAPE} & \multicolumn{1}{c|}{RMSE} & \multicolumn{1}{c}{MAE} & \multicolumn{1}{c}{MAPE} & \multicolumn{1}{c}{RMSE} \\
    \hline
    ARIMA & 31.23 & 19.25 & 33.47 & 4.23  & 5.72  & 7.68  & 33.89 & 17.60  & 46.38 & 34.26 & 21.35 & 44.68 \\
    STResNet & 23.25 &15.58 & 32.25& 4.03&5.68  &7.55 &29.36 & 15.24& 42.46 & 29.25 & 20.33  &40.63 \\
    ACFM &15.86 &10.13& 25.34& 3.52&5.28 &7.86 & 25.86& 11.83& 39.03& 26.38& 19.24& 37.86\\
    STGCN & 17.50  & 11.29 & 27.03 & 3.55  & 5.30   & 7.38  & 25.32 & 11.16 & 39.27 & 25.68 & 18.43 & 36.44 \\
    DCRNN & 17.86 & 11.45 & 27.83 & 3.59  & 4.90   & 7.42  & 25.3  & 11.66 & 38.58 & 25.75 & 18.62 & 36.91 \\
    GWNet & 19.13 & 12.68 & 31.05 & 3.53  & 4.93  & 7.57  & 26.85 & 12.12 & 42.78 & 24.73 & 17.46 & 35.86 \\
    AGCRN & 15.95 & 10.09 & 25.22 & 3.32  & 5.01  & \underline{7.33}  & 22.37 & 9.12  & 36.55 & 25.03 & 17.93 & 36.23 \\
    PDFormer & \textbf{13.58} &\textbf{9.05} & 23.51& 3.46& 4.97& 7.46&\textbf{20.42} &\underline{8.86}& \underline{32.87} &25.22&18.96&35.43\\
    Bi-STAT &\underline{13.62} & \underline{9.43}& \underline{23.17}& 3.42& 4.89 & 7.54& 21.13& 8.98 & 33.86& 25.36& 19.03&35.27 \\
    GPT-ST &14.85 &9.63&24.32 &3.43 &4.97 &7.42& 21.43& 9.32& 34.76& 24.93& 17.52& 35.86 \\
    % STLLM & 16.35	&11.07	&25.96	&3.55	&5.56	&7.72	&23.25	&12.03	&38.46	&26.27	&19.25 &	36.17 \\
    UniST &16.01 & 10.23  &25.37  &3.49  &4.88 &7.51& 24.62& 11.56& 38.21& 25.53& 18.87 &35.46 \\
    AGS   & 15.50  & 9.71  & 25.01 & \textbf{3.15}  & \textbf{4.56}  & 7.52  & 21.56 & 9.03  & 34.90  & 24.96 & 17.88 & 36.02 \\
    BigST & 16.35 & 10.05 & 24.37 & 3.63  & 5.13  & 7.76  & 22.86 & 10.02 & 34.03 & \underline{23.36} & \underline{16.81} & \underline{34.98} \\
    GSNet(Ours)  & 14.76 & 9.48  & \textbf{22.52} & \underline{3.20}   & \underline{4.82}  & \textbf{6.60}   & \underline{20.70}  & \textbf{8.77}  & \textbf{32.66} & \textbf{19.76} & \textbf{14.39} & \textbf{30.99} \\

    \hline
    \end{tabular}%
    
  \label{table:comparison1}%
\end{table*}%

\subsection{Baselines}
We compare the proposed method in this paper with current state-of-the-art approaches in the field. 
The selected baselines are categorized into five groups based on their model frameworks. 
The first category includes classical statistical methods, specifically ARIMA~\cite{williams2003modeling}. 
The second category consists of CNN-based approaches, such as STResNet~\cite{Zhang2017Deep}, ACFM~\cite{Liu2019ACFMAD} and STGCN~\cite{yu2018spatio}. 
The third category focuses on GNN-based methods, with some incorporating RNN structures, represented by DCRNN~\cite{li2018dcrnn}, GWNet~\cite{wu2019graph}, AGCRN~\cite{bai2020adaptive}, AGS~\cite{kdd2023ags} and BigST~\cite{bigst2024}. 
The fourth category includes transformer-based methods, denoted by PDFormer~\cite{pdformer} and Bi-STAT~\cite{bistat}. 
Lastly, the fifth category includes large model (pretrained neural networks) approaches, denoted as GPT-ST~\cite{li2023gptst} and UniST~\cite{Unist}.
:

\begin{itemize}
    \item ARIMA: Historical Average, which models the traffic as a seasonal process and uses the average of previous seasons (e.g., the same time slot of previous days) as the prediction;
    \item STResNet: Spatio-Temporal Residual Network, which combines both spatial and temporal dependencies to forecast crowd movement patterns in urban environments based on residual learning;
    \item ACFM: Attention-based Contextual Feature Mapping, which leverages both spatial and temporal information to predict traffic patterns;
    \item STGCN: Spatio-Temporal Graph Convolutional Networks, which equationte the problem on graphs and build the model with complete convolutional structures;
    \item DCRNN: Diffusion Convolution Recurrent Neural Network, which combines graph convolution with recurrent neural networks in an encoder-decoder manner;
    \item GWNet:  Graph WaveNet is a framework that incorporates adaptive adjacency matrix into graph convolution with 1D dilated convolution;
    \item AGCRN: Adaptive Graph Convolutional Recurrent Network which capture fine-grained spatial and temporal correlations in traffic series automatically based on graph neural networks and recurrent networks;
    \item AGS: Adaptive Graph Sparsification, which propose a graph sparsification algorithm in inference process;
    \item BigST: A linear complexity spatio-temporal graph neural network to efficiently exploit long-range spatio-temporal dependencies for large-scale traffic forecasting;
    \item PDFormer: The proposed PDFormer model introduces a new approach for traffic flow prediction using a transformer model that explicitly accounts for propagation delays in traffic data;
    \item Bi-STAT: Bidirectional Spatial-Temporal Adaptive Transformer, which introduces an innovative architecture that captures both spatial and temporal dependencies in traffic data, while also adapting dynamically to varying traffic conditions;
    \item GPT-ST: A novel approach for improving spatio-temporal graph neural networks through generative pre-training, which uses a generative pre-training strategy to learn effective representations of spatio-temporal data before fine-tuning the model for specific prediction tasks;
    % \item \textcolor{brown}{STLLM: Spatio-Temporal Graph Learning with Large Language Model,which focuses on exploring the capacity of Large Language Models to handle the dynamic nature of spatio-temporal data in urban systems;}
    \item UniST: An universal architecture utilizes prompt-based learning to guide the model's attention to relevant spatial and temporal features, allowing it to adapt efficiently to various urban prediction scenarios.

\end{itemize}

\subsection{Experimental Setup}
Our implementation is based on python 3.12.1, torch 2.2.0, and numpy 1.26.4. 
tested on Ubuntu 20.04.3 (CPU: Intel(R) Xeon(R) CPU E5-2620 v4 @ 2.10GHz, GPU:  GeForce RTX 3080 10GB). 
Tests on PEMS07, PEMS08 and CA, we adopt 60 minutes as the history time window, a.k.a. 12 observed data points ($S$ = 12) are used to forecast traffic in the next 60 minutes ($T$ = 12).
Tests on England, we adopt 180 minutes as the history time window, a.k.a. 12 observed data points ($S$ = 12) and forecast the traffic in the next 180 minutes ($T$ = 12).
In our model, both the Feature Extractor and Relational Compressor are implemented as three-layer stacked structures.
Adam optimizer is used for training our model with an initial learning rate of 0.0001. 
The best parameters are chosen through a carefully parameter-tuning process on the validation set.
The parameters used in the baselines are set to the default values from the original author's publicly available code. 
On certain datasets, due to the increased data scale, we adjusted parameters such as the batch size to ensure the model could run without memory constraints.
PEMS07 and PEMS08 are split by ratio of 7:1:2 in chronological order while England and CA are split by ratio of 6:2:2.
For all datasets, the missing values are filled by the linear interpolation. 
In our model, the values for $C$, $d$, and $C'$ are set to be equal. 
The batch size for the input data is set to 16. Should we encounter issues with insufficient GPU memory, we will reduce the batch size of the model incrementally until the model can operate without such constraints.
We adopt three evaluation metrics commonly used in the field of traffic forecasting to evaluate the performance of models:
Mean Absolute Errors (MAE) represents the average absolute value of prediction errors. For each data point, the absolute error between the actual and predicted values is calculated, and then the average is taken across all data points.
\begin{align}
	MAE = \frac{1}{TN}\sum_{i=1}^{i=T}\sum_{j=1}^{j=N}|\hat{X}^{(t+i)}_j - X^{(t+i)}_j|
\end{align}
Here $\hat{X}$ denotes output of models.
Mean Absolute Percentage Errors (MAPE) measures the percentage of prediction errors. It calculates the relative error between the actual and predicted values for each data point, takes the average across all data points, and expresses the result as a percentage.
\begin{align}
	MAPE = \frac{100\%}{TN}\sum_{i=1}^{i=T}\sum_{j=1}^{j=N}|\frac{\hat{X}^{(t+i)}_j - X^{(t+i)}_j}{X^{(t+i)}_j}|
\end{align}
Root Mean Squared Errors (RMSE) is the square root of the average squared difference between the actual and predicted values for each data point, providing an overall measure of prediction error.
\begin{align}
	RMSE = \sqrt{\frac{1}{TN}\sum_{i=1}^{i=T}\sum_{j=1}^{j=N}(\hat{X}^{(t+i)}_j - X^{(t+i)}_j)^2}
\end{align}

\begin{table*}[t]
\centering

\caption{Computation cost during training and inference on different models.}
    \renewcommand{\arraystretch}{1.1}
    \setlength{\tabcolsep}{7.5pt}

\begin{tabular}{l|cccc|cccccc}
\hline
     & \multicolumn{4}{c|}{Training Time(s/epoch)}       & \multicolumn{4}{c}{Inference Time(s/epoch)} \\
         \hline
        Model  & PEMS08    & England   & PEMS07    & CA    & PEMS08    & England   & PEMS07    & CA \\

    \hline
    STResNet & 32.35 &48.32& 44.17& 153.35& 5.93& 8.21& 7.02 &23.26\\
    ACFM &35.44 & 51.32&  62.37& 172.64& 6.53& 8.74& 6.96& 27.64 \\
    GWNet & 114.23 & 130.75 & 235.32 & 1757.35 &10.35  & 14.86 & 12.12 & 437.53 \\
    AGCRN & 121.35 & 147.36 & 264.16 & 1921.66 & 11.72 & 16.74 & 14.35 & 463.23 \\
    PDFormer &270.25 &325.37& 446.36& 2653.82& 23.42& 35.28& 29.64& 653.24 \\
    Bi-STAT & 160.34 & 203.25& 352.33& 2325.24& 15.37& 20.03& 19.86& 502.51\\
    GPT-ST & 125.32& 150.25& 242.37& 1869.32& 12.01 & 16.07 &15.33 & 467.53 \\
    % STLLM &155.74	&186.32	&302.85	&2351.22	&13.51	&18.04	&15.50	&511.71 \\
    UniST & 703.25& 859.36& 1560.21& 6358.25& 36.21& 40.35& 38.39& 1036.72 \\
    AGS   & 135.74 & 153.25 & 282.37 & 2135.12 & 12.13 & 16.93 & 14.76 & 470.03 \\
    BigST & \underline{27.95} & \underline{42.52} & \underline{37.28} & \underline{95.72} &  \underline{5.86} & \underline{6.75}  & \underline{3.27}  & \underline{16.37} \\
    GSNet(Ours)  & \textbf{10.37} & \textbf{17.62}  & \textbf{11.80} & \textbf{27.26} & \textbf{0.99}  & \textbf{2.12}  & \textbf{1.21}  & \textbf{5.50} \\
    
    Improvement &  2.70$\times$ & 2.41$\times$ & 3.16$\times$  &3.51$\times$ & 5.91$\times$ & 3.18$\times$&  4.84$\times$&  2.98$\times$  \\

    \hline
    \end{tabular}%
    
\label{table:comparison2}
\end{table*}

\begin{figure}
  \centering
  \includegraphics[width=\linewidth]{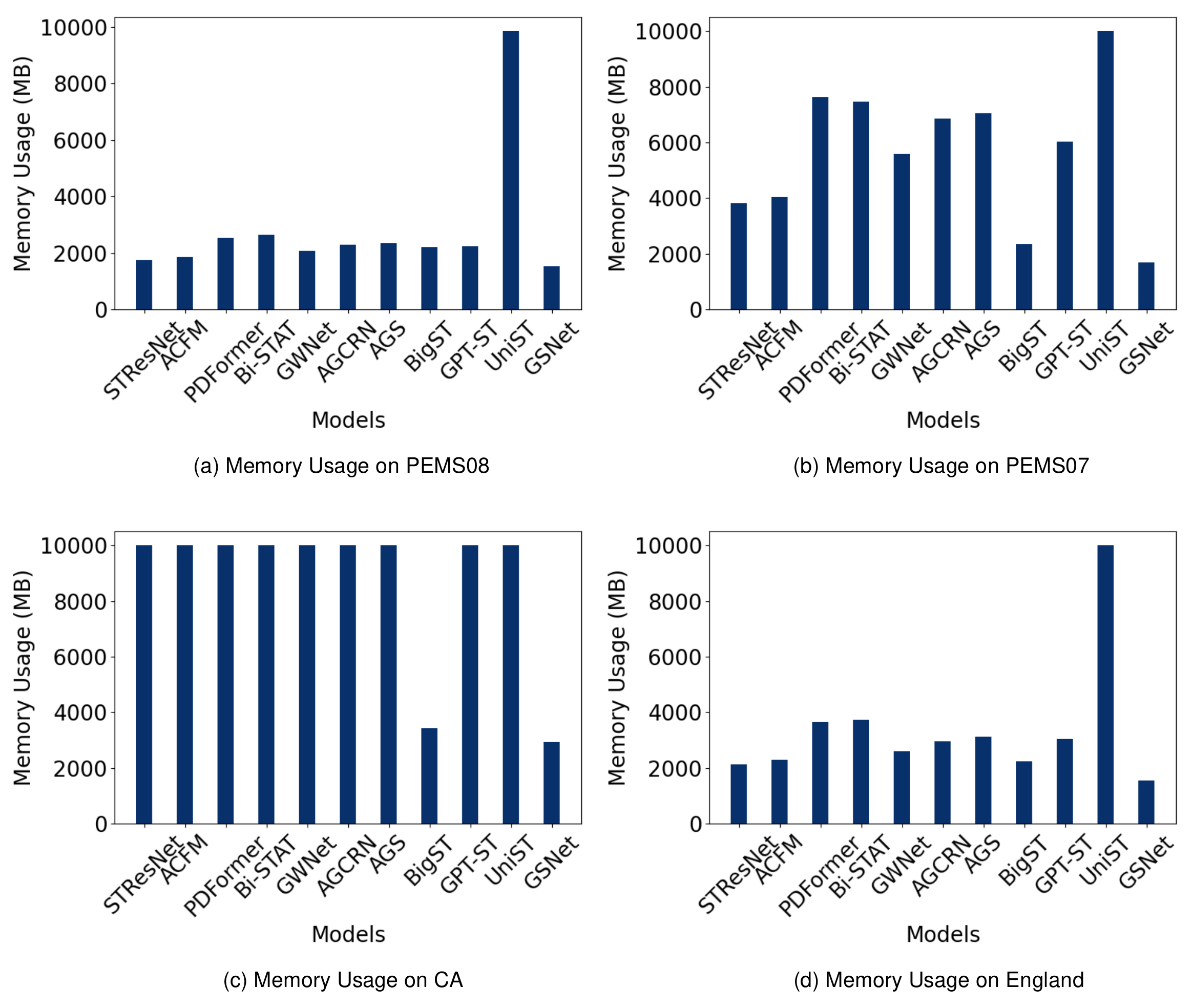}
  \caption{Memory usage on all the datasets.}
  \label{fig:memusage}
\end{figure}

\subsection{Comparison with Existing Models}

Tables~\ref{table:comparison1},~\ref{table:comparison2} and Figure~\ref{fig:memusage} present the effectiveness and efficiency of multiple models across various datasets.

\subsubsection{Performance Comparison}
As shown in the table~\ref{table:comparison1}, our method demonstrates strong predictive performance across multiple datasets. Notably, on larger-scale datasets such as CA, it achieves the best results across all metrics. 
On the PEMS07 and England dataset, our approach consistently achieves either the best or second-best performance across various evaluation metrics. 
Additionally, On the PEMS08 dataset, our method outperforms others specifically in the RMSE metric.
Traditional statistical methods such as ARIMA are not effective in handling complex spatio-temporal data and often exhibit poor performance, whereas graph convolution methods significantly enhance model accuracy.
Methods based on CNN, such as STResNet and ACFM, utilize deep learning techniques. 
Compared to traditional statistical methods, these approaches capture nonlinear features and significantly improve the model's prediction accuracy. 
However, the design framework of CNNs inherently limits their ability to consider adjacency relationships in a more detailed manner. 
Specifically, CNNs are restricted to using a regular receptive field, which makes it challenging to account for higher-order relationships between different nodes in a road network. 
As a result, there is still room for improvement in prediction accuracy for these methods.
DCRNN and STGCN improve upon GCN by employing bidirectional propagation processes, achieving good results compared to traditional statistical methods. 
GWNet and AGCRN optimizes the generation of adaptive adjacency matrices, leading to better performance than other graph neural network methods. 
AGS integrates the advantages of multiple spatio-temporal graph neural network methods, resulting in improved inference accuracy compared to other approaches. 
This method achieves the best accuracy on some metrics in the England dataset and performs slightly less optimally on the PEMS07 and PEMS08 datasets. 
BigST is designed with scalability in mind for graph neural networks; its proposed linear graph convolution offers high scalability, allowing the model to perform well on larger datasets (e.g., CA), but its predictive accuracy on smaller datasets is not as competitive as other methods.
Transformer-based methods, as seen in PDFormer and Bi-STAT, have shown promising performance in prediction tasks. Compared to other approaches, the attention mechanism in Transformers excels at capturing temporal information within time series data, leading to significant improvements in prediction accuracy. However, the complexity of the model design limits its scalability. As a result, these methods struggle to maintain high prediction accuracy when applied to large-scale datasets.
GPT-ST improves the training effectiveness through a pretraining strategy. 
Compared to other GNN-based methods, it significantly enhances the model's prediction accuracy.
On the other hand, UniST is designed with a large model framework that supports training on multiple types of data. 
Its main advantage lies in its ability to perform zero-shot and few-shot learning. 
However, in the experiments presented in this paper, we maintained fairness in comparison by not utilizing multiple data types for training, which resulted in less favorable performance.

\subsubsection{Efficiency Comparison}
Figure~\ref{fig:memusage} illustrates the memory usage of various methods. 
It is evident that GSNet and BigST are the best and second-best performers, respectively. 
Additionally, as the number of nodes in the dataset increases, the advantage of the linear space complexity of GSNet and BigST becomes increasingly apparent.
Apart from GSNet and BigST, all other methods exhibit a cache occupancy of 10GB on the CA dataset. 
Furthermore, UniST consistently consumed 10GB of memory cache across all datasets except for PEMS08.
This uniformity is attributed to the excessive number of nodes in the dataset, which leads to an 'out of memory' error when attempting to run the programs normally. 
Consequently, we reduced the batch size of them incrementally until it could be executed under the condition of a 10GB GPU memory constraint.

Table~\ref{table:comparison2} lists the computational efficiency of various graph neural network models, including training and inference times. 
Our method and BigST achieve the best and second-best runtime results across all datasets.
Although AGS performs sparsification on the adaptive matrices in the model, the original paper and code do not provide methods for efficient computation using these sparse matrices, resulting in longer training and inference time compared to other methods. 
STResNet and ACFM demonstrate competitive efficiency, ranking just behind the BigST model. 
As CNN-based methods, they do not need to account for adjacency relationships between all nodes, which contributes to their relatively fast runtime.
GWNet, AGCRN and GPT-ST do not show a significant difference in runtime compared to our method on smaller datasets (e.g., PEMS08), but their computational times increase dramatically as the number of nodes in the dataset grows. 
In contrast, PDFormer and Bi-STAT require significantly longer training and inference times. This is due to the substantial computational overhead inherent in transformer-based methods. Similarly, UniST exhibits even greater computational demands, with both its training and inference times far exceeding those of other models, primarily due to its complex framework design.

On the CA dataset, GSNet demonstrates exceptional training efficiency. Specifically, it achieves a training acceleration that is 64 to 70 $\times$ faster compared to GWNet and AGCRN.
BigST, despite designing a graph convolution operation with a complexity of $O(N)$, includes a kernel function based on random mapping, which makes the overall operational efficiency of the model slightly lower than ours. 
On the CA dataset, our method achieves 3.51 $\times$ higher training acceleration.
And On the PEMS07 dataset, our method achieves 4.84 $\times$ higher inference acceleration.

\begin{figure*}[!t]
  \centering
  \includegraphics[width=\linewidth]{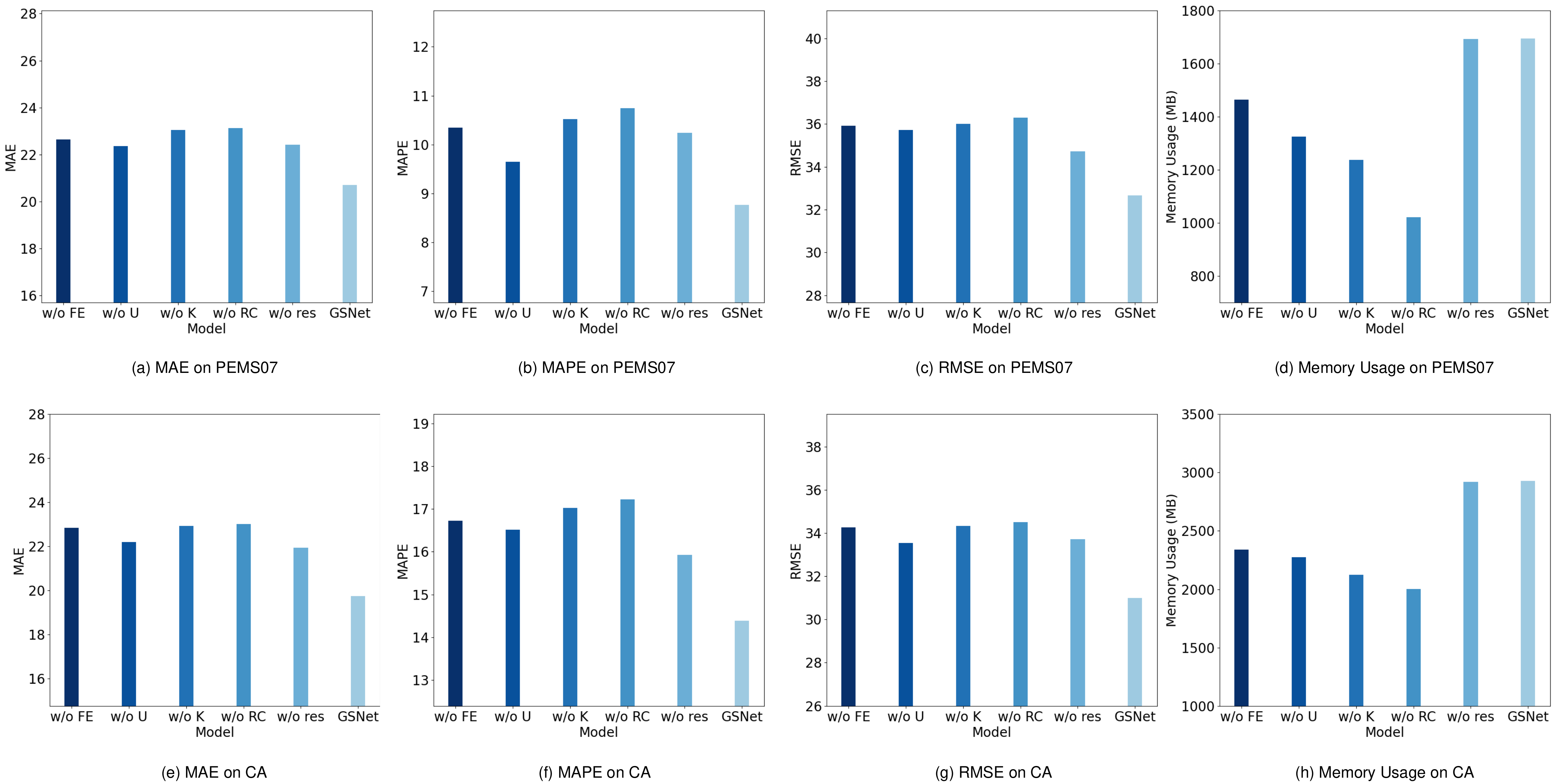}
  \caption{Prediction performance of different variants on PEMS07 and CA datasets.}
  \label{fig:ab}
\end{figure*}

In summary, among existing graph neural network methods, some methods, such as GWNet and AGCRN, sacrifice model scalability for higher predictive accuracy, leading to a decline in both accuracy and efficiency on large-scale datasets. 
Other methods, like BigST enhance model scalability at the expense of predictive accuracy.
Our method balances the effectiveness and efficiency of the model, achieving the scalability of high-accuracy graph neural networks on large-scale datasets.

\begin{figure}
  \centering
  \includegraphics[width=\linewidth]{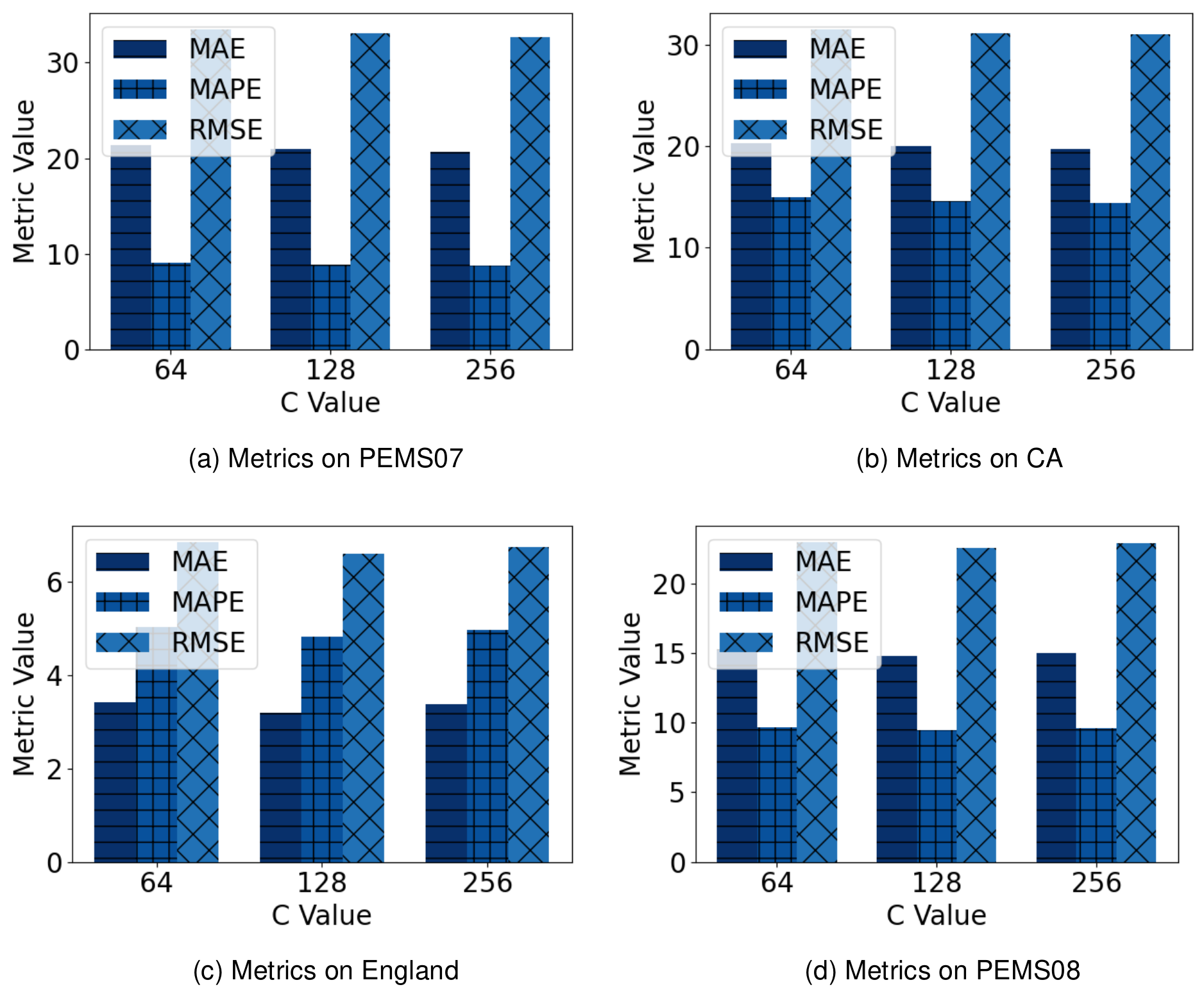}
  \caption{Prediction performance of different C value on all the datasets.}
  \label{fig:Cvalue}
\end{figure}

\subsection{Ablation Studies}

To further verify effectiveness of the proposed method, we conduct ablation studies on several datasets. 
In particular, we design five variants of our model:

(1) $w/o~FE$: This variant removes Feature Extractor from the model, constructing the model using only Relational Compressor.

(2) $w/o~U$: This variant removes the source concatenation from Relational Compressor but retains the adjacency matrix within Relational Compressor.
The matrix $K$ has also been appropriately adjusted in terms of its shape to ensure that the matrix multiplication is well-defined.

(3) $w/o~K$: This variant removes the adjacency matrix from Relational Compressor while retaining the source concatenation.
The adjacency matrix $K$ has been replaced with an identity matrix of the same shape.

(4) $w/o~RC$: This variant removes Relational Compressor, retaining only the Feature Extractor.

(5) $w/o~res$: This variant removes the residual connections from both Feature Extractor and Relational Compressor.

As illustrated in Figure~\ref{fig:ab}, we first compared the predictive accuracy of different variants across multiple datasets. 
It can be observed that removing Relational Compressor or the adjacency matrix within Relational Compressor has the most significant impact on the model's accuracy. 
This indicates that the fusion of graph node features is a crucial step in enhancing predictive accuracy. 
Removing Feature Extractor from the model also greatly affects accuracy, demonstrating that accurately learning node features is a prerequisite for accurate prediction.
Removing the residual connections from the model significantly reduces its accuracy as well. 
This is because Feature Extractor and Relational Compressor are combined in a sequential manner, and residual connections effectively mitigate the vanishing or exploding gradient issues in the stacked structure. 
It is noteworthy that almost all variants result in a substantial decrease in the model's predictive accuracy, highlighting the collaborative relationship between different modules in our approach and validating the effectiveness of our overall model framework design.

We also compared the memory usage of different variants. 
It can be observed that the variant $w/o~RC$ has the most significant reduction in memory usage. 
This indicates that the Relational Compressor in our model consumes a considerable amount of memory usage, whereas the Feature Extractor, due to its simpler structure, uses less memory.

\begin{figure}
  \centering
  \includegraphics[width=\linewidth]{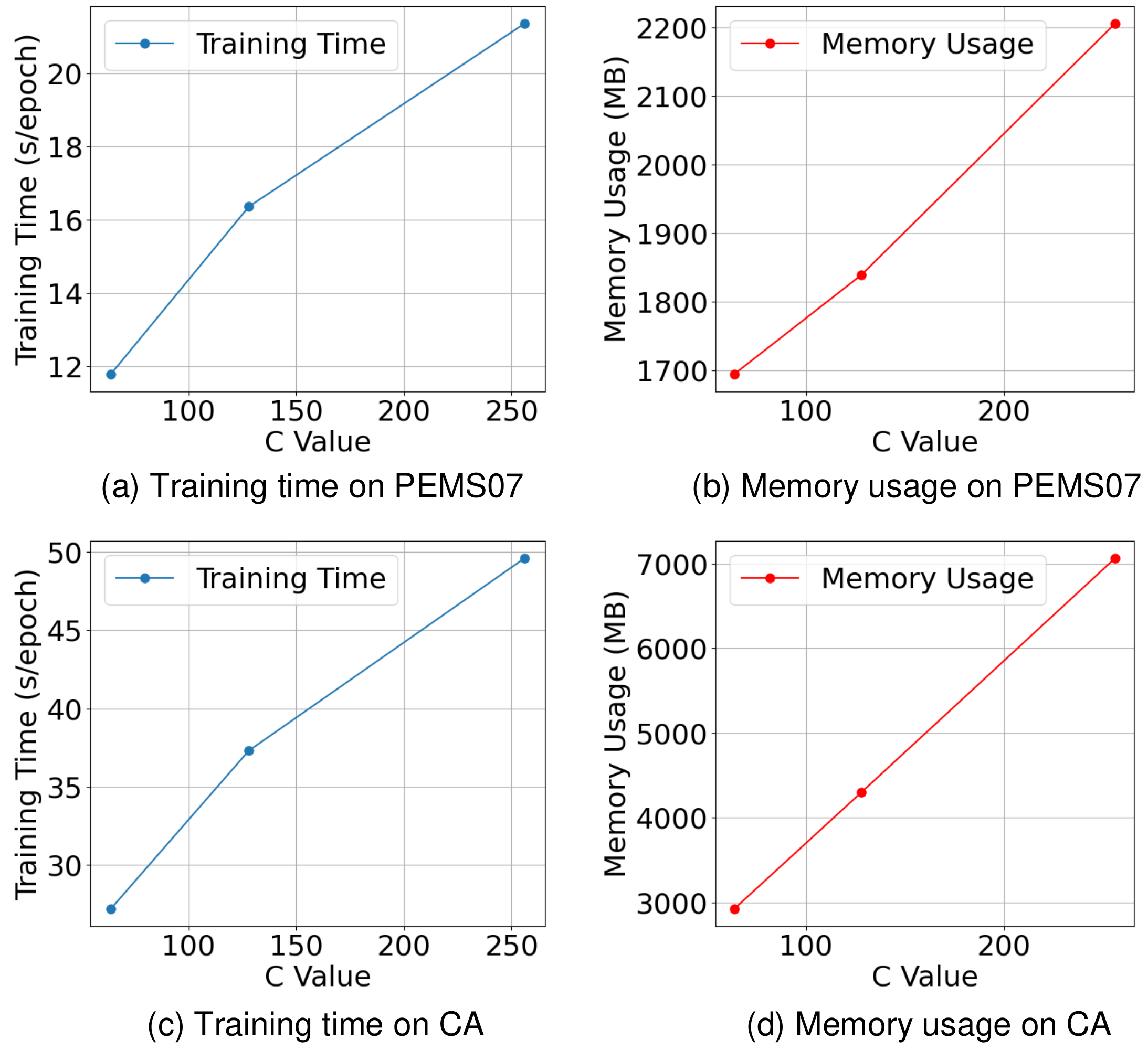}
  \caption{Memory usage and training time of different C value on PEMS07 and CA datasets.}
  \label{fig:hiddim}
\end{figure}

\begin{figure}
  \centering
  \includegraphics[width=\linewidth]{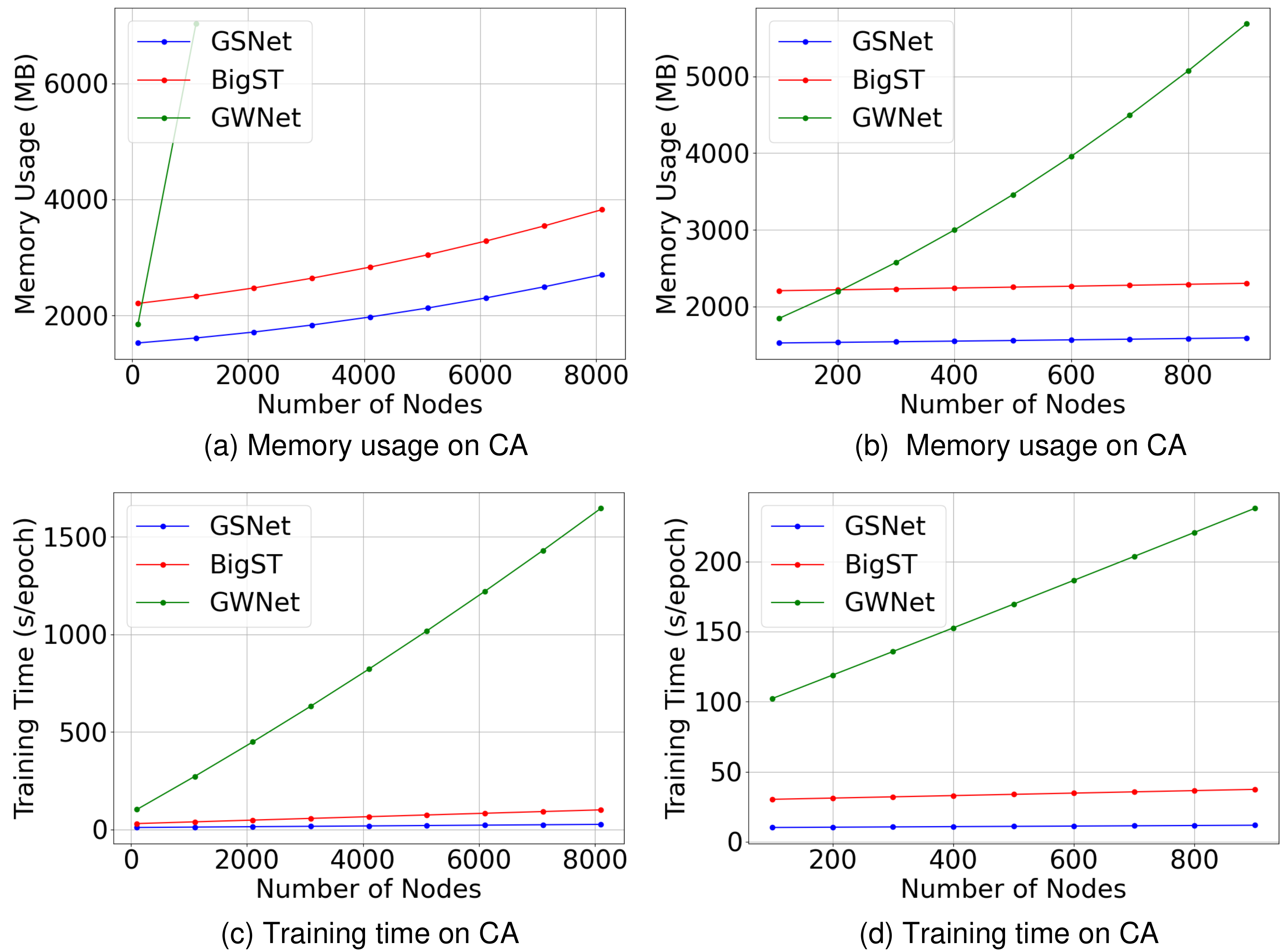}
  \caption{Memory usage and training time of different models with varying N.}
  \label{fig:memtime}
\end{figure}

\begin{figure}
  \centering
  \includegraphics[width=\linewidth]{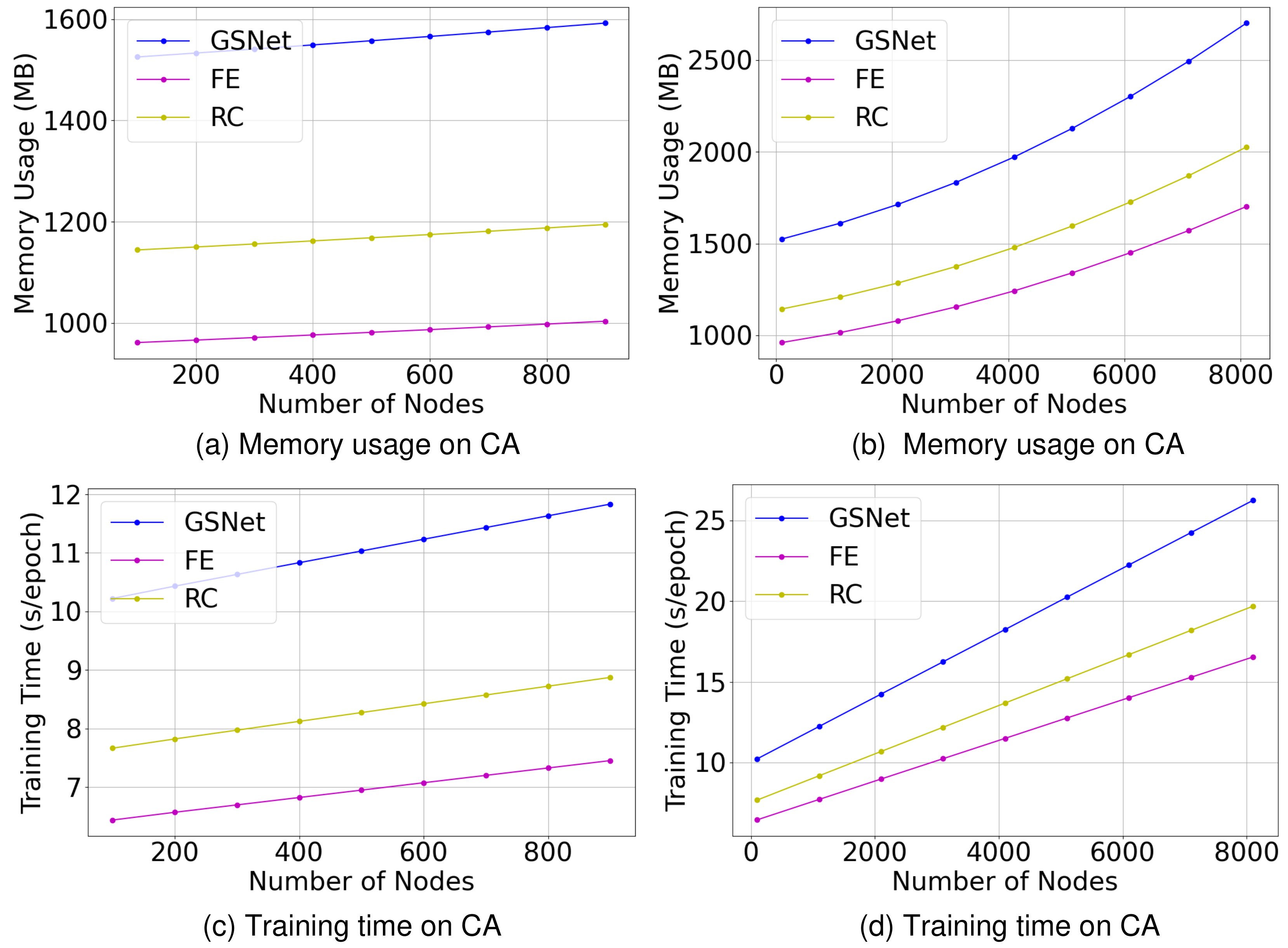}
  \caption{Memory usage and training time of modules in the GSNet with varying N.}
  \label{fig:memtimenew}
\end{figure}
\subsection{Parameter Sensitivity}

In the model, we compress the input data into a low-dimensional space to model the implicit associations between nodes, where the dimension of $K$ is also a hyperparameter. 
As shown in Figure~\ref{fig:Cvalue}, we conducted a series of experiments to test the model's performance under different values of $C$. 
It can be seen that on the PEMS08 dataset, the model's accuracy first increases and then decreases with the increase of $C$. 
On the PEMS07 and CA datasets, the model's predictive accuracy increases with the increase of $C$. 
We speculate that on larger datasets, the model's performance might also decrease with the increase of $C$, but due to equipment limitations, we have not tested this hypothesis on larger values of $C$. 
The experimental results indicate that for large-scale datasets, appropriately increasing the value of $C$ can enhance the model's predictive accuracy.
As illustrated in Figure~\ref{fig:hiddim}, we also tested the model's operational efficiency and memory usage under different values of $C$. 
It can be observed that as the value of $C$ increases, the model's training time, inference time, and memory usage also increase. 
The increase in training and inference time is relatively slow, while memory usage almost linearly increases with the increase of $C$.
In summary, when using GSNet on larger-scale datasets, the value of $C$ can serve as a trade-off coefficient between effectiveness and efficiency, adjusting the balance between the model's predictive accuracy and operational efficiency.

\subsection{Efficiency Analysis}

To further assess the computational efficiency of our model across different data scales, we segmented the CA dataset based on sensor ID order and compared our method against several baselines on these varying scales. 
As shown in Figures~\ref{fig:memtime}, both the running time and memory usage of our method incrementally increase with the dataset size. 
Notably, the rate of this increase is more gradual compared to existing methods, demonstrating our model's superior scalability.
With GWNet, we observed that as the number of nodes surpasses 1,000, it consumes the entire memory capacity of our GPU. 
To mitigate this, we adapted the model to utilize a smaller batch size, thereby reducing the volume of data processed per batch. 
It is evident that, among methods with linear complexity, GraphSparseNet (GSNet) significantly outperforms BigST in terms of storage requirements and training efficiency.
This comprehensive analysis underscores the robustness and adaptability of our model, offering valuable insights into its performance and efficiency across a spectrum of data scales and settings.

We further examine the efficiency of the two core modules, the Feature Extractor and Relational Compressor, as the data scale increases.
To do so, we experiment with two model variants: FE, which retains only the Feature Extractor module, and RC, which retains only the Relational Compressor module.
As illustrated in the figure~\ref{fig:memtimenew}, both modules exhibit a roughly linear increase in spatio-temporal complexity with respect to the number of nodes. 
It is worth noting that the Relational Compressor includes an additional feature fusion process, which leads to higher training time and increased memory usage.

\section{Related Work}

Traffic flow forecasting stands as a quintessential challenge in spatio-temporal data prediction, with analogous tasks including the forecasting of shared bicycle demand, as well as the demand for buses and taxis, and the prediction of crowd flows, among others \cite{Li2015bike, Chai2018bike, hu2021bus, zhao2019bus}. Traditional statistical approaches such as ARIMA \cite{williams2003modeling} and SVM \cite{1997svm}, while prevalent in time series forecasting, often fall short due to their inability to account for spatial dimensions, making them less effective for complex spatio-temporal data sets.
The advent of deep learning has introduced methods adept at handling the intricacies and non-linearities inherent in traffic data. Convolutional Neural Networks (CNNs), in particular, have become a staple in traffic flow forecasting \cite{Zhang16DNN, Zhang2017Deep, Ouyang2022, yao2018modeling, Liu2019ACFMAD}. These networks interpret traffic flow data as images, where each pixel represents the traffic count within a specific grid cell over a given time interval. By leveraging techniques originally developed for image recognition \cite{Ted2022survey}, CNNs can effectively model the spatial relationships between different grid regions.
Recurrent Neural Networks (RNNs) have also been instrumental in the analysis of sequence data, bringing their sequence memorization capabilities to bear on traffic flow forecasting \cite{ye2019kdd, shi2015nips, zon2018ijcai}. 
More recent advancements have seen Graph Neural Networks (GNNs) rise to prominence for their ability to manage the spatio-temporal correlations present in traffic flow data, achieving state-of-the-art results \cite{Pan2022meta, Shen2022ttpnet, Sun2022multiview, Guo2022LearningD}. GNNs, initially designed for graph structure analysis, have found widespread application in node embedding \cite{pan2018} and node classification \cite{kipf2016semi}.
In the realm of transportation systems, GNNs, including graph convolutional and graph attention networks, have been adapted to model graph structures and have achieved remarkable performance. For instance, DCRNN \cite{li2018dcrnn} employs a bidirectional diffusion process to emulate real-world road conditions and utilizes gated recurrent units to capture temporal dynamics. ASTGCN \cite{guo2019attention}, on the other hand, utilizes dual attention layers to discern the dynamics of spatial dependencies and temporal correlations.
STGCN, Graph WaveNet, LSGCN, and AGCRN \cite{yu2018spatio, wu2019graph, huang2020lsgcn, bai2020adaptive} represent a lineage of methods that build upon Graph Convolutional Networks (GCNs) to extract spatio-temporal information. Notably, Graph WaveNet introduces a self-adaptive matrix to factor in the influence between nodes and their neighbors, while LSGCN employs an attention layer to achieve a similar end.
STSGCN, STFGNN, and STGODE \cite{song2020stsgcn, li2021stfgnn, zheng2021ode} propose GCN methodologies designed to capture spatio-temporal information in a synchronous manner. MTGNN \cite{Wu2020MTGNN} introduces a graph learning module that constructs a dynamic graph by calculating the similarity between learnable node embeddings. DMSTGCN \cite{Han2021DMSTGCN} captures spatio-temporal characteristics by forging dynamic associations between nodes.
STPGNN \cite{STPGNN} enhances predictive accuracy by taking into account special nodes within the road network. 
In addition to GNN-based methods, several approaches leveraging Transformers \cite{pdformer, bistat} and pre-training \cite{li2023gptst, Unist} have also been proposed. These methods have also shown promising results. However, both Transformer-based and pre-training approaches, along with GNNs, still encounter significant computational challenges when identifying nodes relationships, which limits the scalability of these models on large-scale datasets.
Existing adaptive graph neural network methods often rely on fully connected graphs to bolster the model's learning capabilities. Yet, the exponential growth in the number of edges with an increase in nodes poses a challenge for these methods when generalizing to larger-scale datasets.
To counter this, AGS \cite{kdd2023ags} has proposed a method for significantly simplifying the adaptive matrix, thereby reducing the model's computational load. However, this approach is limited to the inference stage. In practical applications, the computational cost during the training phase often dwarfs that of the inference phase. A method that maintains linear complexity during training can significantly enhance the model's operational efficiency.
BigST \cite{bigst2024} introduces a method that employs kernel functions to linearly approximate graph convolution operations, yielding a graph prediction model with linear complexity. However, kernel-based methods can sometimes result in anomalous gradient values during training, impacting model convergence and, by extension, the model's performance.
Amidst the ever-expanding scale of traffic data, there is an urgent need for graph neural network methods capable of delivering high-precision predictions at scale.

\section{Conclusion}

In this paper, we addressed the challenge of enhancing the scalability of GNN-based methods for large-scale traffic spatio-temporal data prediction. We identified that existing methods either do not fully address the computational complexity of graph operations or compromise model accuracy when simplifying GNN structures. 
By introducing GraphSparseNet (GSNet), we have provided a novel approach that leverages two module—the Feature Extractor and the Relational Compressor—to effectively manage and analyze large-scale traffic data. 
Our theoretical analysis and empirical evaluations highlight the significant advantages of GSNet over existing methods. By reducing model complexity to \(O(N)\), GSNet addresses the scalability issue while maintaining high predictive accuracy. 
Our extensive experiments on real-world traffic datasets demonstrate that GSNet not only achieves high predictive accuracy but also significantly improves training efficiency, outperforming state-of-the-art methods in both aspects. 
The model’s efficiency is further validated by a substantial improvement in training time, outperforming current linear models by 3.51x.
Future work may explore extending our framework to other spatio-temporal prediction problems beyond traffic flow, further broadening its applicability and impact in various real-world domains.

%\clearpage

\bibliographystyle{ACM-Reference-Format}
\bibliography{sample}

\end{document}